\documentclass[default,iicol]{sn-jnl}% Default with double column layout

\jyear{2022}%

\theoremstyle{thmstyleone}%
\theoremstyle{thmstyletwo}%

\theoremstyle{thmstylethree}%
\usepackage{overpic}
\usepackage{arydshln}
\usepackage{marvosym}

\newcommand\blfootnote[1]{%
  \begingroup
  \renewcommand\thefootnote{}\footnote{#1}%
  \addtocounter{footnote}{-1}%
  \endgroup
}

\definecolor{myGray}{gray}{.5}

\newcommand{\figref}[1]{Fig.~\ref{#1}}
\newcommand{\tabref}[1]{Table~\ref{#1}}
\newcommand{\secref}[1]{Sec.~\ref{#1}}

\def\ie{\emph{i.e.}}
\def\eg{\emph{e.g.}}
\def\etc{\emph{etc}}

\def\ourmodel{MVLT}

\begin{document}

\title[MVLT]{\textbf{Masked Vision-Language Transformer in Fashion}}

\author[1]{\fnm{Ge-Peng} \sur{Ji}$^\dagger$}
%\equalcont{Contributed equally.}

\author[1]{\fnm{Mingcheng} \sur{Zhuge}$^{\dagger}$}
%\equalcont{Contributed equally.}

\author[1]{\fnm{Dehong} \sur{Gao}}

\author[2]{\fnm{Deng-Ping} \sur{Fan}$^*$}

\author[2]{\fnm{Christos} \sur{Sakaridis}}

\author[2]{\fnm{Luc Van} \sur{Gool}}

% \affil[1]{\orgdiv{Research School of Engineering}, \orgname{Australian National University}, \orgaddress{\city{Canberra 2601}, \country{Australia}}}

% % 
% \affil[2]{\orgdiv{AI Initiative}, \orgname{King Abdullah University of Science and Technology}, \orgaddress{\city{Thuwal 23955}, \country{Saudi Arabia}}}

\affil[1]{\orgdiv{International Core Business Unit}, \orgname{Alibaba Group}, \orgaddress{\city{Hangzhou 310051}, \country{China}}}

% \affil[2]{\orgdiv{School of Cybersecurity}, \orgname{Northwestern Polytechnical University}, \orgaddress{\city{Xi'an 710072}, \country{China}}}
 
\affil[2]{\orgdiv{Computer Vision Lab}, \orgname{ETH Zürich}, \orgaddress{\city{Zürich 8092}, \country{Switzerland}}}

\abstract{
We present a masked vision-language transformer (\textbf{\ourmodel}) for fashion-specific multi-modal representation. Technically, we simply utilize vision transformer architecture for replacing the BERT in the pre-training model, making \ourmodel~the first end-to-end framework for the fashion domain.
Besides, we designed masked image reconstruction (\textbf{MIR}) for a fine-grained understanding of fashion.
\ourmodel~is an extensible and convenient architecture that admits raw multi-modal inputs without extra pre-processing models (\eg, ResNet), \textit{implicitly} modeling the vision-language alignments. More importantly, \ourmodel~can easily generalize to various matching and generative tasks. Experimental results show obvious improvements in retrieval (rank@5: \textbf{17\%}) and recognition (accuracy: \textbf{3\%}) tasks over the Fashion-Gen 2018 winner Kaleido-BERT.
Code is made available at \burl{https://github.com/GewelsJI/MVLT}.
}


\maketitle

%%%%%%%%%%%%%%%%%%  
\section{Introduction}
%%%%%%%%%%%%%%%%%%  

\begin{figure}[t!]
  \centering
  \includegraphics[width=.98\linewidth]{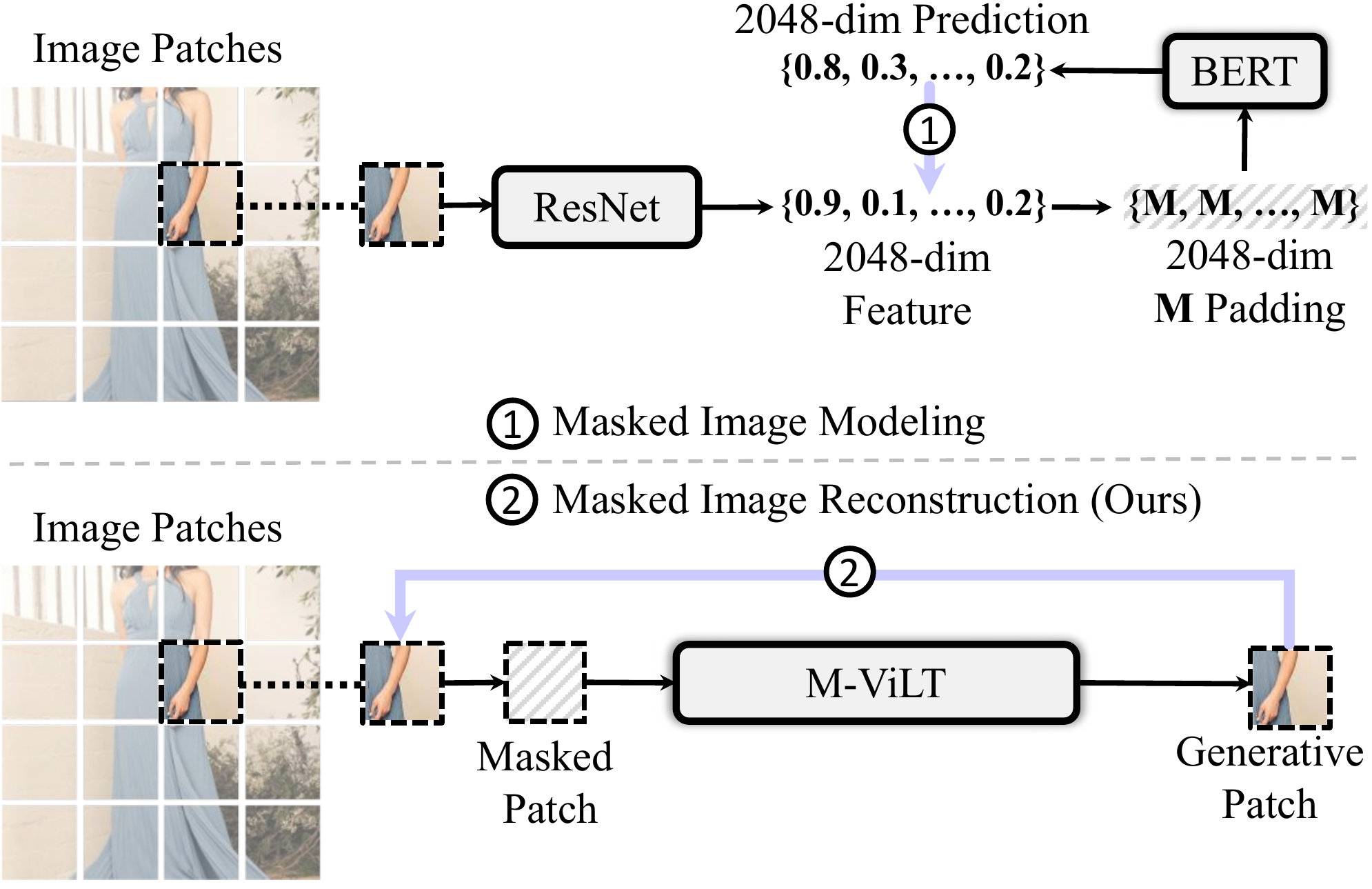}
  \caption{
    Different visual reconstruction tasks for VL pre-training~\cite{gao2020fashionbert,Zhuge2021KaleidoBERT} utilize masked image modeling (top) with the random masking strategy (\ie, to use M padding to replace raw vectors), which reconstructs pre-extracted visual semantics (\ie, probabilities) at the feature-level. 
    We introduce a generative task named masked image reconstruction (bottom), which directly reconstructs image patches at the pixel level.
  }\label{fig:mask_motivation}
\end{figure}

The emergence of transformer\blfootnote{$\dagger$ Contributed equally. $^*$ Corresponding author. Work was done while Ge-Peng Ji was an research intern in Alibaba Group.} is drawing enormous attention from the academic community, facilitating the advancement of computer vision (CV)~\cite{vit2021ICLR,liu2021Swin} and natural language processing (NLP)~\cite{vaswani2017attention,sun2022paradigm}.
Benefiting from the robustness of transformers, researchers also contribute to the vision-language (VL) field~\cite{agarwal2021evaluating,chen2020generative,lin2021m6,ramesh2021zero,wu2021fashion} with zeal. 
To better utilize the pre-trained models in CV and NLP, existing general VL models are mainly based on the BERT model~\cite{devlin2019bert} or adopt the well-pretrained vision extractors~\cite{he2016deep,ren2015faster} or both. 
However, general VL methods~\cite{qi2020imagebert, lu2019vilbert, chen2020uniter} still struggle when applied to the fashion domain in e-commerce because they suffer from the two main issues: 
\textbf{a) Insufficient Granularity.} Unlike the general objects with complex backgrounds, only focusing on coarse-grained semantics is insufficient for a fashion product~\cite{hsiao2019fashion++, vasileva2018learning,fan2022domain}, as it would lead the network to generate sub-optimal results. Contrarily, the fashion-oriented framework requires more fine-grained representations, such as a suit with different materials (\eg, wool, linen, and cotton) or collars (\eg, band, camp, and windsor). 
\textbf{b) Bad Transferability.} The pre-extracted visual features are not discriminative for fashion-oriented tasks, restricting the cross-modal representations.

To address the above issues, we present a novel VL framework, termed masked vision-language transformer (\ourmodel). 
Specifically, we introduce a generative task, masked image reconstruction (MIR), for the fashion-based VL framework.
Compared to previous pre-training tasks, such as masked image modeling (regression task) or masked image classification (classification task), MIR enables the network to learn more fine-grained representations via pixel-level visual knowledge (see \figref{fig:mask_motivation}).
Further, inspired by pyramid vision transformer (PVT)~\cite{wang2021pyramid}, we utilize a pyramid architecture for our VL transformer. Then, we introduce the MIR task. These two improvements significantly enhance the ability to adapt to fashion-specific understanding and generative tasks, and can conduct in an end-to-end manner. 
To this end, \ourmodel~can directly process the raw multi-modal inputs in dense formats (\ie, linguistic tokens and visual patches) without extra (\eg, ResNet) pre-processing models~\cite{yang2020fashion, al2020paris}. 
Our main contributions are summarized as follows:

\begin{itemize}

\item We introduce a novel masked image reconstruction (\textbf{MIR}) task, which is the first real pixel-level generative strategy utilized in VL pre-training.

\item Based on the MIR task, we present an end-to-end VL framework, called \textbf{\ourmodel}, for the fashion domain, greatly promoting the transferability to the downstream tasks and large-scale web applications.

\item Extensive experiments show that \ourmodel~significantly outperforms the state-of-the-art models on matching and generative tasks.

\end{itemize}

%%%%%%%%%%%%%%%%%%  
\section{Background}\label{sec:related_work}
%%%%%%%%%%%%%%%%%%  

\begin{figure*}[t!]
  \centering
  \includegraphics[width=0.9\linewidth]{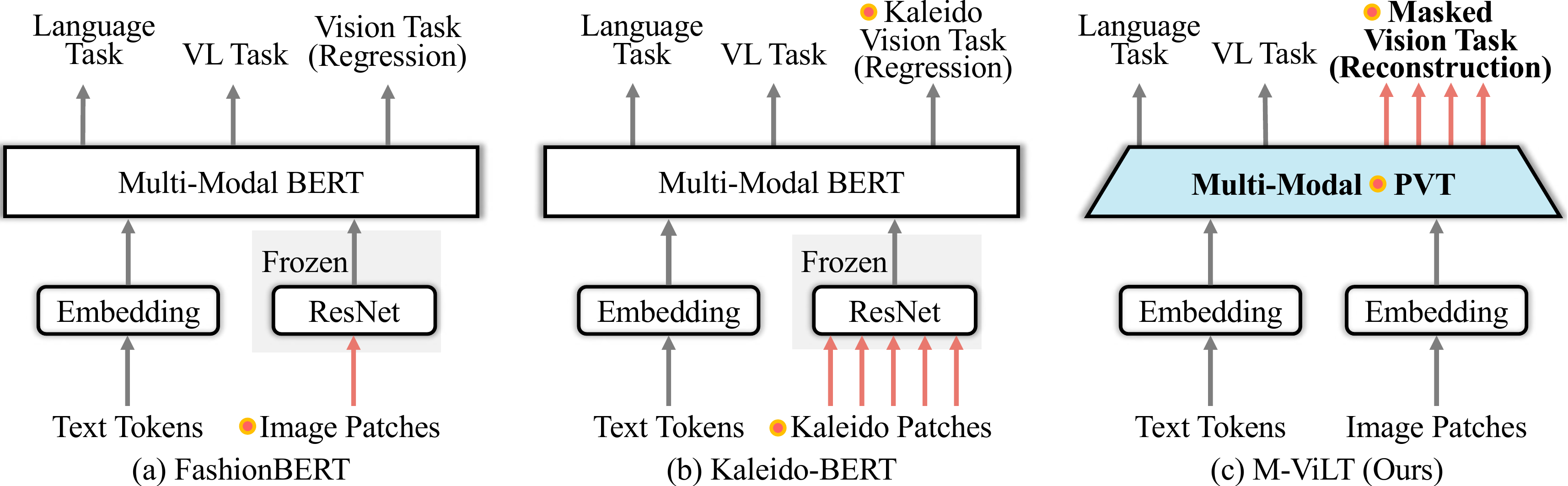}
  \caption{
    Comparison of \ourmodel~to cutting-edge fashion-oriented VL frameworks. FashionBERT (a) utilizes a language-based encoder (\ie, BERT) to extract VL representations with single-scale visual input (\ie, image patches). Kaleido-BERT (b) extends it with two upgrades: adds five fixed-scale inputs (\ie, Kaleido patches) to acquire hierarchical visual features and designs Kaleido vision tasks to fully learn VL representations. However, the visual embedding of these models is frozen (\ie, without parameter updating), thus, a lack of domain-specific visual knowledge severely hinders their transferability. Differently, our \ourmodel~(c) adaptively learns hierarchical features by introducing masked vision tasks in an end-to-end framework, significantly boosting the VL-related understanding and generation.
  }\label{fig:framework_comp}
\end{figure*}

In recent years, BERT-based pre-training models have been widely investigated in VL tasks. Many previous attempts, such as LXMERT~\cite{tan2019lxmert}, VL-BERT~\cite{su2020vl}, and FashionBERT~\cite{gao2020fashionbert}, were successful in a wide range of downstream applications.
Experiments and discussions show that BERT is a powerful method for learning multi-modal representations, outperforming several previous CNN-based~\cite{lee2018stacked} or LSTM-based~\cite{niu2017hierarchical,xia2022skating} approaches.
Compared to previous studies, this paper aims to develop a more efficient self-supervised objective that can be easily implemented in pre-training and provides better representations for real-world applications.
Thus, we review research on masked learning strategies and end-to-end multi-modal schemes that inspired us the most.

%%%%%%%%%%%%%%%%%%  
\subsection{Masked Learning Strategies}
%%%%%%%%%%%%%%%%%%  

Masked modeling is the vital self-supervised task in BERT~\cite{devlin2019bert} and initially demonstrates outstanding abilities in natural language processing. Researchers have replicated its strength in language models because of its utility in multi-modal and vision tasks. Most VL works~\cite{su2020vl, lu2019vilbert, li2020oscar} transfer masked modeling into visual tokens and use a \textit{regression} task to construct the token feature from nonsense-replace or a \textit{classification} task to predict the token's attribute. 
To reduce the difficulty in learning, Kaleido-BERT~\cite{Zhuge2021KaleidoBERT} optimizes masked modeling by employing a Kaleido strategy that facilitates coherent learning for multi-grained semantics. Although this work improves the performance of VL-related tasks in fashion indeed, we argue that the token-patch pre-alignment scheme by using auxiliary tool~\cite{zhuge2021salient, xu2015show}  is still complex and impedes the application to practical settings.
Another work~\cite{arici2021mlim} introduces the MLIM approach that strengthens the masked image modeling with an \textit{image reconstruction} task, which shares a similar idea to ours. However, our experiments showed that requiring a model to reconstruct the entire image without any reminder is too difficult. 
Recently, BEiT~\cite{bao2022beit} and MAE~\cite{he2021masked} utilize a BERT-style pre-training as part of the visual learner, and they discover that models are effective at learning semantics with such a scheme. These two works strengthen our conviction that converting the original masked image modeling (\ie, a regression task) to a masked image reconstruction task is possible. 
However, our primary goal is to design a generative pretext task that makes the multi-modal modeling in VL pre-training easier while eliminating the need for using prior knowledge. It will be extremely helpful in our practical application setting with billion-level data.

%%%%%%%%%%%%%%%%%%  
\subsection{End-To-End Multi-Modal Schemes}
%%%%%%%%%%%%%%%%%%  

Pixel-BERT~\cite{huang2020pixel} is the first method to consider end-to-end pre-training. It employs 2$\times$2 max-pooling layers to reduce the spatial dimension of image features, with each image being downsampled 64 times. Although this work sets a precedent for end-to-end training, such a coarse and rigid method cannot work well in practical settings because it is simply combined with a ResNet~\cite{he2016deep} as part of joint pre-training, without considering the loss in speed and performance. 
Recently, VX2TEXT~\cite{lin2021vx2text} proposes to convert all modalities into language space and then perform end-to-end pre-training using a relaxation scheme. Though it is exciting to translate all the modalities into a unified latent space, it ignores that the usage of data extracted by pre-trained methods as input to the model cannot be regarded as an end-to-end framework.
% it ignores that using raw data (\ie, image, video, and text) as input without pre-trained models isn't a true end-to-end solution.
%  
According to the timeline, ViLT~\cite{kim2021vilt} is the first method that indeed investigates an end-to-end framework via replacing region- or grid-based features with patch-based projections. However, without other designs, it cannot obtain competitive performance since it is just a vanilla extension of ViT~\cite{vit2021ICLR}. 
Grid-VLP~\cite{yan2021grid} is similar to ViLT, but it takes a further step by demonstrating that using a pre-trained CNN network as the visual backbone can improve performance on downstream tasks.
SOHO~\cite{huang2021seeing} takes the entire image as input and creates a visual dictionary to affine the local region. However, this method does not fit fashion-specific applications due to the lack of reliable alignment information. As a result, the vision dictionary may merely learn the location of the background or foreground rather than complex semantics.
FashionVLP~\cite{Goenka2022fashionvlp} uses a feedback strategy to achieve better retrieval performance. In practice, they use the well-pretrained knowledge extracted from ResNet and then model the whole, cropped, and landmark representations. Besides, they adopt Faster-RCNN as an object detector for popping out RoI candidates.
Besides, some works are designed for end-to-end pre-training~\cite{lei2021less,xu2021e2e,akbari2021vatt}, but they are used for specific tasks and are not directly applicable to our research. 

\begin{figure*}[t!]
  \centering
  \includegraphics[width=\linewidth]{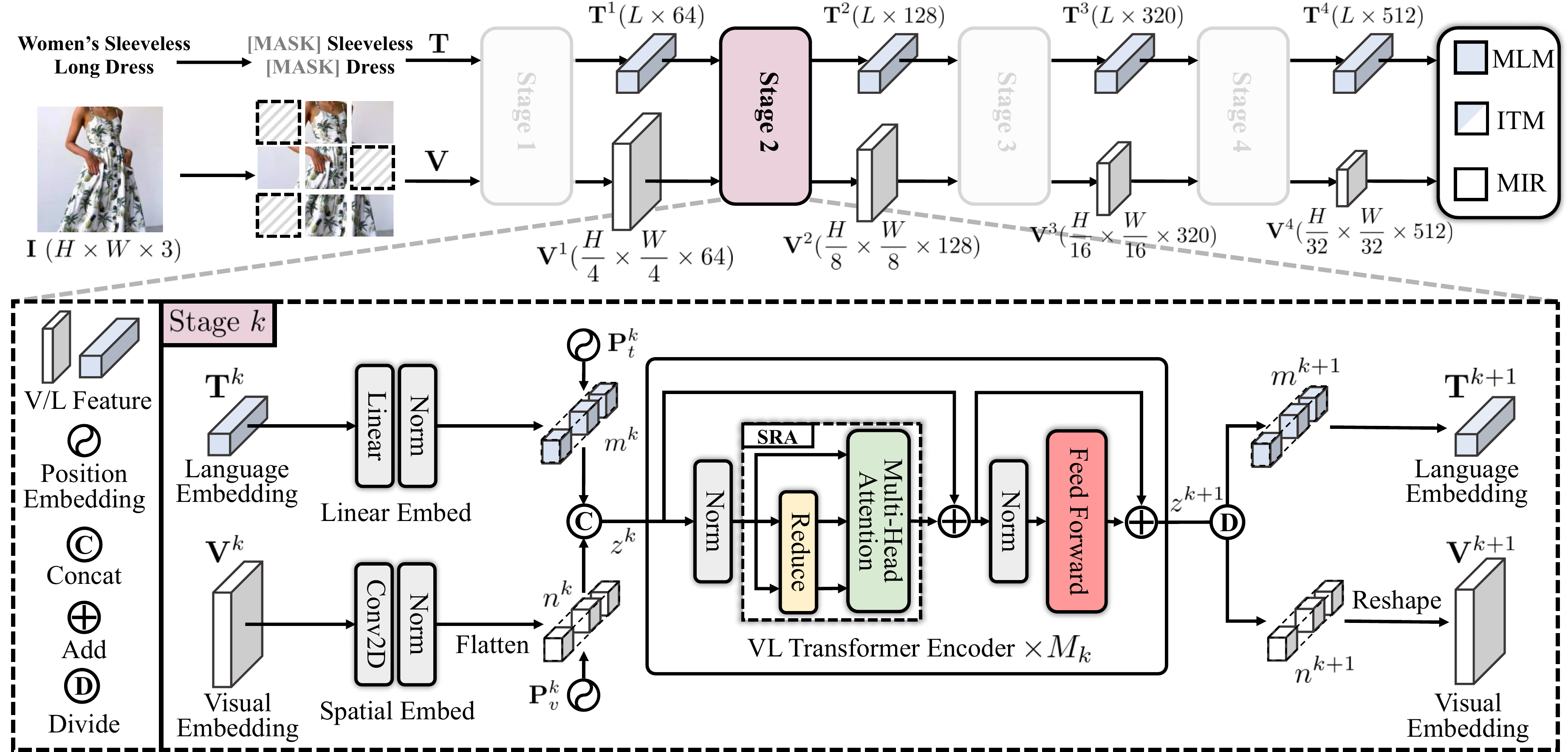}
  \caption{Pipeline of our \ourmodel~framework.
  Our overall architecture consists of four stages containing language and visual embeddings and multiple transformer encoders ($\times M_k$). Introducing the masking strategy for three sub-tasks, \ie, masked image reconstruction (MIR), image-text matching (ITM), and masked language modeling (MLM), our \ourmodel~can be trained in an end-to-end manner. More details can be found in~\secref{sec:M-ViLT}.
  }\label{fig:pipeline}
\end{figure*}

Despite existing methods employing different approaches to construct an end-to-end scheme, solutions that forgo pre-trained methods (\eg, ResNet, BERT) and use raw data (\ie, text, image) as inputs remain under-explored and are needed urgently in multi-modal applications. 

\noindent \textbf{Remarks.} As shown in \figref{fig:framework_comp}, similar to the existing two fashion-based approaches, \ie, FashionBERT (a) and Kaleido-BERT (b), the proposed \ourmodel~(c) is also a patch-based VL learner, which extends the 
pyramid vision transformer~\cite{wang2021pyramid} to an architecture that adaptively extracts hierarchical representations for fashion cross-modal tasks. 
It is the first model that solves the end-to-end problem of VL pre-training in fashion, which allows us to simplify the implementation of our \ourmodel~in the fashion industry using a twin-tower architecture~\cite{yi2019sampling}.

%%%%%%%%%%%%%%%%%%  
\section{Masked Vision-Language Transformer}\label{sec:M-ViLT}
%%%%%%%%%%%%%%%%%%  

Our goal is to build an end-to-end VL framework for the fashion domain. 
The overall pipeline of our \ourmodel~is depicted in \figref{fig:pipeline}. Like PVT, our architecture inherits four stages' properties and generates features with different sizes. Two keys of the proposed architecture are the multi-modal encoder (\secref{sec:multi_modal_encoder}) and the pre-training objectives (\secref{sec:pre_training_loss}).

%%%%%%%%%%%%%%%%%%  
\subsection{Multi-Modal Encoder}\label{sec:multi_modal_encoder}
%%%%%%%%%%%%%%%%%%  

As shown in~\figref{fig:pipeline}, \ourmodel~admits visual and verbal inputs. On the language side, we first tokenize the caption of a fashion product and use the specific token \texttt{[MASK]} to randomly mask out the caption tokens with the masking ratio\footnote{We follow the default setting in BERT~\cite{devlin2019bert}.} $r_l$. Following the masking procedure, we obtain a sequence of word tokens. Then, we insert a specific \texttt{[CLS]} token at the head of this sequence. Besides, we pad the sequence to a unified length $L$ using the \texttt{[PAD]} token if the length is shorter than 128. This procedure generates the language input ids $\mathbf{T} \in \mathbb{R}^{L} = \langle t_1;\cdots; t_{L} \rangle$. On the vision side, we treat $\mathbf{I} \in \mathbb{R}^{H \times W \times 3}$ as visual input, where $H$ and $W$ denote the height and width of the given input. This input is sliced into multiple grid-like patches $\mathbf{V} \in \mathbb{R}^{N \times P \times P \times 3} = \langle v_1; \cdots ; v_N \rangle $, where $N=\frac{HW}{P^2}$ is the total number of patches and $P$ denotes the patch size. Similarly, the split patches are masked out with mask ratio $r_v$. We provide more details about the above masking strategy for the language and vision parts in~\secref{sec:pre_training_loss}.

The above multi-modal inputs are embedded and fed into the consequent four VL interaction stages (\ie, $k \in \{1,2,3,4\}$). In the first stage, we generate the vision and language embeddings, $\mathbf{T}^1$ and $\mathbf{V}^1$ respectively, via the given inputs ($\mathbf{T}$ and $\mathbf{V}$). Regarding the subsequent stages, we consider only the $k$-th stage, to have concise illustrations. As shown in the bottom part of~\figref{fig:pipeline}, we first embed the language embedding $\mathbf{T}^k \in \mathbb{R}^{L \times D_k}$ into the language hidden feature $m^k \in \mathbb{R}^{L \times D_{k+1}}$, which is formulated as:
\begin{equation}
    m^k = \mathbf{T}^k \ast \mathbf{W}^{k}_{t} + \mathbf{P}^{k}_{t},
\end{equation}
where $\mathbf{W}^{k}_{t} \in \mathbb{R}^{D_{k} \times D_{k+1}}$ and $\mathbf{P}^{k}_{t} \in \mathbb{R}^{L \times D_{k+1}} $ are the learnable linear embedding and position embedding matrices. $D_k$ is the size of the hidden feature embedding.

\begin{table}[t!]
	\centering
	\tiny
	\caption{Hyperparameter of our multi-modal encoders.}
	\label{tab:hyper_parameter}
	\renewcommand{\arraystretch}{1.5}
	\renewcommand{\tabcolsep}{3.2mm}
	\begin{tabular}{l|c|c|c|c}
    \hline
    \textbf{Hyperparameter} &$k=1$ &$k=2$ &$k=3$ &$k=4$ \\
    \hline
    Layer number $M_k$  &2 &2 &2 &2 \\\
    Hidden size $D_k$  &64 &128 &320 &512 \\
    Reduction size $R_k$  &$4$ &$8$ &$16$ &$32$ \\
    Kernel size $K_k$  &4 &2 &2 &2 \\
    Stride length $S_k$  &4 &2 &2 &2 \\ 
	\hline
	\end{tabular}
\end{table}

The visual embeddings are $\mathbf{V}^k \in \mathbb{R}^{\frac{H}{R_{k}} \times \frac{W}{R_{k}} \times D_k}$, where $R_{k}$ denotes the spatial reduction factor of visual embedding. To acquire pyramid visual features, $\mathbf{V}^k$ are then embedded and flattened into the visual hidden feature $n^k \in \mathbb{R}^{ (HW/R_{k+1}^2) \times D_{k+1}}$ via a two-dimensional projection (\ie, Conv2D block). In particular, this projection enforces the network to reduce the equivalent spatial dimension from $\mathbb{R}^{HW/R_{k}^2}$ to $\mathbb{R}^{HW/R_{k+1}^2}$ by utilizing the convolutional kernel $\mathbf{W}^k_v \in \mathbb{R}^{D_{k} \times K_k \times K_k \times D_{k+1} }$ with kernel size $K_k$ and stride length $S_k$. This could be formulated as:
\begin{equation}
    n^{k} = \mathbf{Flatten}(\mathbf{V}^k \ast \mathbf{W}^{k}_{v}) + \mathbf{P}^{k}_{v},
\end{equation}
where $\mathbf{P}^{k}_{v} \in \mathbb{R}^{N \times D_{k+1}} $ denotes the position embedding matrix. We then concatenate these two VL hidden features $z^k = \langle m^k ; n^{k} \rangle $ and feed them into multiple ($M_k$) VL transformer encoders. Each encoder contains the multi-head self-attention layer with spatial reduction (\ie, reduce block), multi-layer perceptron, and layer normalization. Finally, we obtain the encoded multi-modal feature $z^{k+1} = \langle m^{k+1} ; n^{k+1} \rangle $ and divide it into a language part $\mathbf{T}^{k+1} = m^{k+1}$ and a visual part $\mathbf{V}^{k+1} = \mathbf{Reshape}(n^{k+1})$, where the $\mathbf{Reshape}(\cdot)$ operation consists in recovering the spatial dimension of the given feature.

After four VL interaction stages, we generate the four text embeddings $\{ \mathbf{T}^k \}_{k=1}^4$ and four pyramid vision embeddings $\{ \mathbf{V}^k \}_{k=1}^4$, respectively.
\tabref{tab:hyper_parameter} presents more detailed hyperparameter settings of our method.

\subsection{Pre-Training Objectives}\label{sec:pre_training_loss}
%%DP Fan
To acquire discriminative multi-modal representations, we adopt three pre-training tasks to establish the inter-and intra-relationships between the most primitive VL modalities, including vision (masked image reconstruction, MIR), language (\ie, masked language modeling, MLM), and VL (image-text matching, ITM) modalities.

\begin{figure}[t!]
  \centering
  \includegraphics[width=\linewidth]{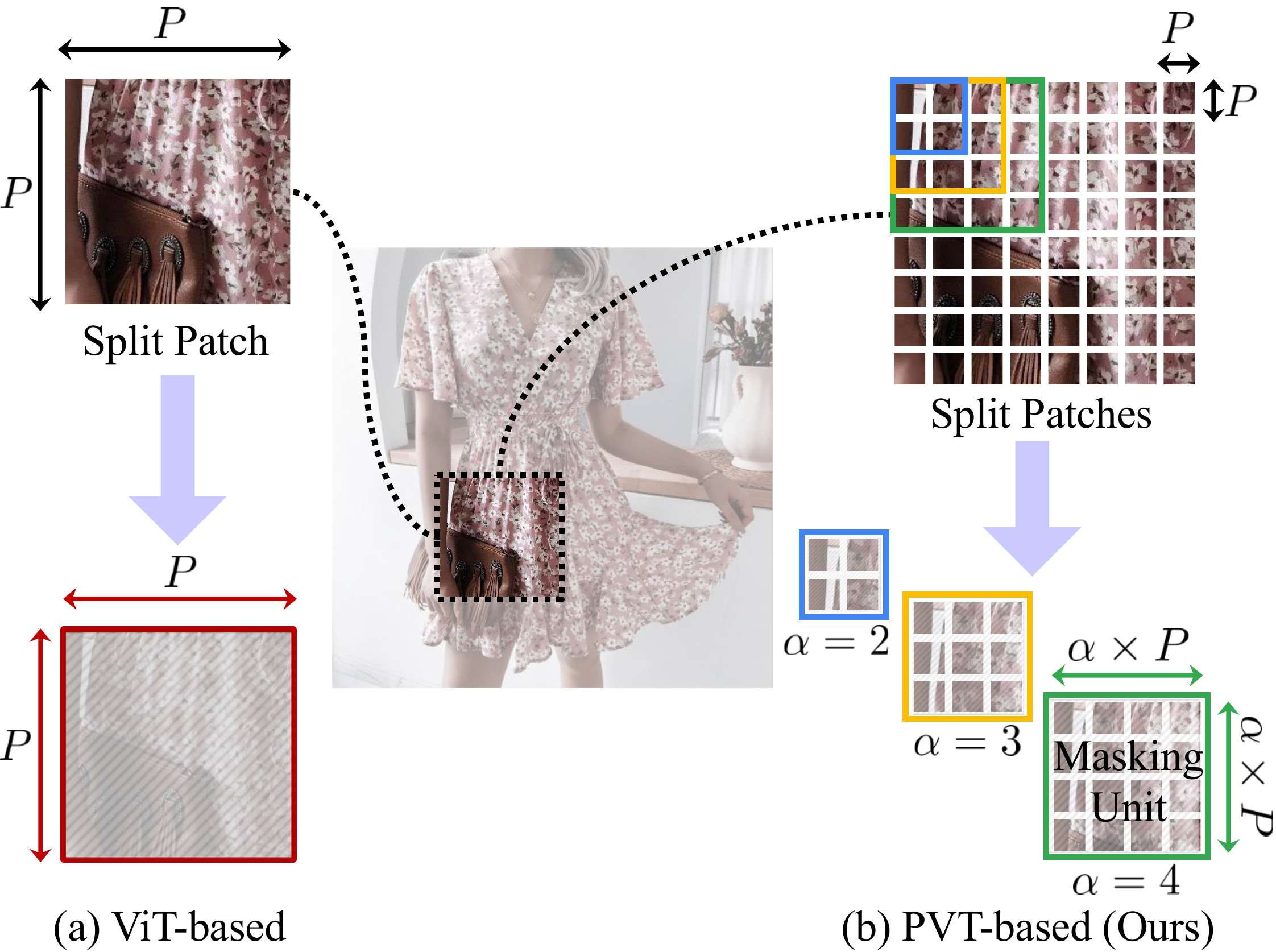}
  \caption{
   %%DP Fan
   PVT-based architectures offer more options for designing the masking strategy. 
   %The double arrow lines denote the height/width of split patch or masking unit. 
   The vanilla ViT-based method (a)~\cite{kim2021vilt} only selects a fixed-scale patch to mask, \ie, $P^{2}$. However, PVT-based method (b) is more versatile because it combines more fine-grained patches as a basic masking unit, \ie, $(\alpha \times P)^2$, where $\alpha \in \{1,2,..,8\}$. These masked patches are not overlapped with each other. This characteristic provides a flexible way to learn the suitable semantics by using different values for $\alpha$. Notably, we adopt a fixed scale factor of masking units in an individual experiment.
  }\label{fig:mask_strategy}
\end{figure}

\noindent{\textbf{Objective 1: Masked Image Reconstruction (MIR).}}
As for the general domain, models are enough to learn the coarse-grained semantics from the patch- or region-based objectives and achieve satisfactory results. However, the fashion-specific models require more fine-grained representations, such as a suit with different materials (\eg, wool) or collars (\eg, Windsor), which needs a pixel-to-pixel vision pre-training objective.
Inspired by the masked language modeling~\cite{devlin2019bert}, we attempt to build the pixel-to-pixel relationships from the perspective of generative tasks, which promote the scalability of visual representations. We design the Masked Image Reconstruction (MIR) to accomplish this idea.
To help our model learn better by MIR, we utilize the pyramid characteristic of PVT architecture~\cite{wang2021pyramid} to design a flexible masking strategy. Unlike the ViT-based method (a) in~\figref{fig:mask_strategy}, PVT-based architecture (b) masks out the input image according to the masking unit matrix that contains small-grained patches. Given the patch sequence $\mathbf{V} = \{ v_n \}_{n=1}^N \in \mathbb{R}^{N \times P \times P \times 3}$, the masked-out sequence $\mathbf{V}_{\backslash \Phi}$ is defined as:
\begin{equation}
\begin{aligned}
    \mathbf{V}_{\backslash \Phi} &= \mathcal{F}_{M} ( \{\mathbf{M}(q;\alpha;\Phi)\}_q^Q, \{v_n\}_{n=1}^N) \\
    &= \begin{cases}
    \texttt{[ZERO]}, & \mathbf{M}(q;\alpha;\Phi) = 1, \\
    v_n, & \mathbf{M}(q;\alpha; \Phi) = 0,
    \end{cases}
\end{aligned}
\end{equation}
where $\mathcal{F}_M(\cdot;\cdot)$ represents a function (or procedure) of our masking strategy, $q$ is the randomly selected area of the masking unit, and \texttt{[ZERO]} means that we use a pixel value of zero\footnote{In fact, we set $\texttt{[ZERO]}=10^{-6}$ to bring better optimization stability and less pattern degradation.} to fill the selected areas.
The masking units $ \{\mathbf{M}(q;\alpha;\Phi)\}_{q=1}^{Q}$ are derived from the indicator function:
\begin{equation}
    \mathbf{M}(q;\alpha;\Phi)= \mathbf{1}(q) = \begin{cases}
    1, & q \in \Phi, \\
    0, & q \notin \Phi,
    \end{cases}
\end{equation}
where each value in a set of integers $\Phi$ is randomly selected from range $[1,Q]$ with ratio $r_v$. $Q=\frac{H \times W}{(\alpha \times P)^2}$ is the total number of masking units.
For instance in~\figref{fig:mask_strategy}~(b), we can define $\alpha$ from 1 to 8. 
In our default settings, we set $\alpha=4$ to capture more fine-grained semantics\footnote{The vanilla masking strategy in \figref{fig:mask_strategy} (a) with $P = 32$ becomes a special case of our masking strategy in \figref{fig:mask_strategy} (b) when $\alpha=8, P = 4$.}.

Since the smooth-$\ell 1$ loss is less sensitive to the outliers, we use it as the pre-training objective to reconstruct the whole image via the masked-out sequence $\mathbf{V}_{\backslash \Phi}$. It is defined as:
\begin{equation}
\scriptscriptstyle{
    \mathcal{L}_{\text{MIR}} = \begin{cases}
    0.5 \times ( \mathbf{I}'_{(x,y)} - \mathbf{I}_{(x,y)})^2,&\text{if}~\mathbf{I}'_{(x,y)}-\mathbf{I}_{(x,y)} < 1,\\
    \mid \mathbf{I}'_{(x,y)}-\mathbf{I}_{(x,y)} \mid - 0.5,&\text{otherwise},
    \end{cases}
}
\end{equation}
where $\mathbf{I}'_{(x,y)}$ and $\mathbf{I}_{(x,y)}$ denote the pixel at coordinate $(x,y)$ in the reconstructed image $\mathbf{I}'$ and the input image $\mathbf{I}$, respectively. $\mathbf{I}' = \mathcal{F}_{\text{MIR}}(\mathbf{V}_{\backslash \Phi};\mathbf{W}_{\text{MIR}})$ is parameterized by learnable weights $\mathbf{W}_{\text{MIR}}$. Function $\mathcal{F}_{\text{MIR}}(\cdot;~\mathbf{W}_{\text{MIR}})$ denotes a standard four-level U-Net~\cite{ronneberger2015u} decoder, which admits four pyramidal vision embeddings $\{ \mathbf{V}^k \}_{k=1}^4$ as inputs.

\noindent{\textbf{Objective 2: Image-Text Matching (ITM).}}
The appended classification embedding in the last language embedding $\mathbf{T}^4$ is used to couple the representations from VL modalities. We utilize the function $\mathcal{F}_{\text{ITM}}(\cdot;\mathbf{W}_{\text{ITM}})$ to denote a full-connected (FC) and softmax layers, parameterized by the weights $\mathbf{W}_{\text{ITM}}$. $\mathcal{F}_{\text{ITM}}$ outputs a two-class probability vector $\mathbf{p}_{\text{ITM}}=\mathcal{F}_{\text{ITM}}(\langle \mathbf{T},\mathbf{V} \rangle ;\mathbf{W}_{\text{ITM}})$, representing whether the input fashion image and caption match (\ie, positive pair) or not (\ie, negative pair). The positive pairs are selected from the same fashion product category, whereas the negative pairs are chosen at random from different entries. The binary cross-entropy loss function finally constrains this task:
\begin{equation}
\begin{aligned}
\mathcal{L}_{\text{ITM}} =& -\mathbb{E}_{\langle \mathbf{T},\mathbf{V} \rangle} [ \mathbf{y}_{\text{ITM}} \log(\mathbf{p}_{\text{ITM}}) \\
&+ (1 - \mathbf{y}_{\text{ITM}}) \log(1 - \mathbf{p}_{\text{ITM}}) ],
\end{aligned}
\end{equation}
where $ \mathbf{y}_{\text{ITM}} $ denotes the ground-truth label, \ie, $1$ for matched pairs and $0$ for unmatched pairs.

\noindent{\textbf{Objective 3: Masked Language Modeling (MLM).}}
Following~\cite{alberti2019fusion}, we randomly use the specific token \texttt{[MASK]} to replace the original text tokens. The target of the MLM is to predict the text content for the masked tokens using the unmasked tokens and patches. Given a tokenized sequence $\mathbf{T} = \{t_1, \dots,t_L\}$, the masked-out sequence is denoted by $\mathbf{T}_{\backslash i} = \{t_1,\dots, \texttt{[MASK]}_i,\dots,t_L\}$. We use the cross-entropy loss to model this objective:
\begin{equation}
    \mathcal{L}_{\text{MLM}} = -\mathbb{E}_{\mathbf{T}}[\log(\mathbf{p}_{\text{MLM}})],
\end{equation}
where $\mathbf{p}_{\text{MLM}}=\mathcal{F}_{\text{MLM}}( \mathbf{T}_{\backslash i}; \mathbf{W}_{\text{MLM}} )$ denotes the predicted probability for each masked-out token $\texttt{[MASK]}_i$ using $\mathbf{T}_{\backslash i}$. The function $\mathcal{F}_{\text{MLM}}(\cdot;\mathbf{W}_{\text{MLM}})$ represents the parameters $\mathbf{W}_{\text{MLM}}$ of a classifier. The final pre-training objective of the proposed \ourmodel~is a combination of the three objectives:
\begin{equation}
    \mathcal{L}_{\text{total}} = w_{1}\times\mathcal{L}_{\text{MIR}} + w_{2}\times\mathcal{L}_{\text{ITM}} + w_{3}\times\mathcal{L}_{\text{MLM}}.
\end{equation}

\begin{table*}[t!]
\centering
\tiny
\renewcommand\arraystretch{1.3}
\setlength\tabcolsep{0.8pt}
\caption{
Retrieval (\ie, TIR and ITR) and recognition (\ie, M-CR and S-CR) performances on Fashion-Gen dataset. $\uparrow$ means the larger, the better. Here, $\text{Sum}\mathcal{R}$=($\mathcal{R}$@1+$\mathcal{R}$@5+$\mathcal{R}$@10) $\times 100$ and $\text{Sum}\mathcal{C}$=$(\mathcal{A}+\text{macro}$-$\mathcal{F}) \times 100$. ``N/A'' means the score is not available. ``Diff'' means the numerical difference between the performance of the second-ranked competitor and our \ourmodel.}
\label{tab:exp_i2t_t2i}
    \begin{tabular}{c|lc||cccccccccc|cc}
        \hline
        &&& \textbf{VSE} & \textbf{VSE++} & \textbf{SCAN} & \textbf{PFAN} & \textbf{ViLBERT} & \textbf{ImageBERT} & \textbf{FashionBERT} & \textbf{VL-BERT} & \textbf{OSCAR} & \textbf{Kaleido-BERT} &\multicolumn{2}{c}{\textbf{\ourmodel}} \\
        Task &\multicolumn{2}{l||}{Metric} &arXiv$_{14}$ &BMVC$_{18}$ &ECCV$_{18}$ &arXiv$_{19}$ &NeurIPS$_{19}$ &arXiv$_{20}$ &SIGIR$_{20}$ &ICLR$_{20}$ &ECCV$_{20}$ &CVPR$_{21}$ & \textbf{OUR$_{22}$} & Diff \\
        \hline
         \multirow{4}{*}{TIR}
         & $\mathcal{R}$@1 & $\uparrow$ & 4.350\% & 4.600\% & 4.300\% & 6.200\% & 21.12\% & 24.78\% & 26.75\% & 22.63\% & 25.10\% &\underline{33.88\%} & \textbf{34.60\%} & \textcolor{myGray}{$\scriptscriptstyle +0.72\%$} \\
         & $\mathcal{R}$@5 & $\uparrow$ & 12.76\% & 16.89\% & 13.00\% & 20.79\% & 37.23\% & 45.20\% & 46.48\% & 36.48\% & 49.14\% & \underline{60.60\%} & \textbf{78.00\%} &\textcolor{myGray}{$\scriptscriptstyle +17.40\%$}  \\
         & $\mathcal{R}$@10 & $\uparrow$ & 20.91\% & 28.99\% & 22.30\% & 31.52\% & 50.11\% & 55.90\% & 55.74\% & 48.52\% & 56.68\% & \underline{68.59\%} &\textbf{89.50\%} &\textcolor{myGray}{$\scriptscriptstyle +20.91\%$} \\  
        &$\text{Sum}\mathcal{R}$& $\uparrow$ &38.02	&50.48	&39.6	&58.51	&108.46	&125.88	&128.97	&107.63	 &130.92 &\underline{163.07} & \textbf{202.1} & \textcolor{myGray}{$\scriptscriptstyle +39.03$} \\
         \hline
         \multirow{4}{*}{ITR}
         & $\mathcal{R}$@1 & $\uparrow$ & 4.010\% & 4.590\% & 4.590\% & 4.290\% & 20.97\% & 22.76\% & 23.96\% & 19.26\% & 23.39\% & \underline{27.99\%} & \textbf{33.10\%} & \textcolor{myGray}{$\scriptscriptstyle +5.11\%$}  \\
         & $\mathcal{R}$@5 & $\uparrow$ & 11.03\% & 14.99\% & 16.50\% & 14.90\% & 40.49\% & 41.89\% & 46.31\% & 39.90\% & 44.67\% & \underline{60.09\%} &\textbf{77.20\%} &\textcolor{myGray}{$\scriptscriptstyle +17.11\%$} \\
         & $\mathcal{R}$@10 & $\uparrow$ & 22.14\% & 24.10\% & 26.60\% & 24.20\% & 48.21\% & 50.77\% & 52.12\% & 46.05\% & 52.55\% & \underline{68.37\%} & \textbf{91.10\%} &\textcolor{myGray}{$\scriptscriptstyle +22.73\%$} \\
        &$\text{Sum}\mathcal{R}$& $\uparrow$ 
         &37.18	&43.68	&47.69	&43.39	&109.67	&115.42	&122.39	&105.21	&120.61	&\underline{156.45} & \textbf{201.4} & \textcolor{myGray}{$\scriptscriptstyle +44.95$} \\
         \hline
         \multirow{3}{*}{M-CR}
         &$\mathcal{A}$ & $\uparrow$ 
         &N/A &N/A&N/A&N/A &N/A &90.77\% &91.25\% &N/A &91.79\% &\underline{95.07\%} &\textbf{98.26\%} &\textcolor{myGray}{$\scriptscriptstyle +3.19\%$}\\
         &$\text{macro}$-$\mathcal{F}$ & $\uparrow$ 
         &N/A &N/A&N/A&N/A &N/A &0.699 &0.705 &N/A &\underline{0.727} &0.714 &\textbf{0.896} &\textcolor{myGray}{$\scriptscriptstyle +0.169$} \\
         &$\text{Sum}\mathcal{C}$ & $\uparrow$ 
         &N/A &N/A&N/A&N/A &N/A &160.67 &161.75 &N/A &164.49 &\underline{166.47} &\textbf{187.86} &\textcolor{myGray}{$\scriptscriptstyle +21.39$} \\
         \hline
         \multirow{3}{*}{S-CR}
         &$\mathcal{A}$ & $\uparrow$ 
         &N/A &N/A&N/A&N/A &N/A &80.11\% &85.27\% &N/A &84.23\%&\underline{88.07\%} &\textbf{93.57\%} &\textcolor{myGray}{$\scriptscriptstyle +5.50\%$} \\
         &$\text{macro}$-$\mathcal{F}$ & $\uparrow$ 
         &N/A &N/A&N/A&N/A &N/A &0.575 &0.620 &N/A &0.591 &\underline{0.636} &\textbf{0.829} &\textcolor{myGray}{$\scriptscriptstyle +0.193$} \\
         &$\text{Sum}\mathcal{C}$ & $\uparrow$ 
         &N/A &N/A&N/A&N/A &N/A &137.61 &147.27 &N/A &143.33 &\underline{151.67} &\textbf{176.47} &\textcolor{myGray}{$\scriptscriptstyle +24.80$}\\
    \hline
    \end{tabular}
\end{table*}

%%%%%%%%%%%%%%%%%%  
\subsection{Downstream Tasks}\label{sec:downstream_app}
%%%%%%%%%%%%%%%%%%  

For a fair comparison, we follow the same training/inference protocols as in~\cite{gao2020fashionbert,Zhuge2021KaleidoBERT} and also adopt the Fashion-Gen 2018~\cite{rostamzadeh2018fashion} benchmark as the base of our experiments. This dataset contains $67,666$ fashion products (\ie, $60,147$ entries for training and $7,519$ entries for testing) and their associated product descriptions. Each product corresponds to an image set (including $1 \sim 6$ samples) at various viewing angles. As a result, we utilize $260,480$ and $35,528$ image-text pairs as training and testing partitions, respectively. For a fair comparison, we test \ourmodel~and compared models on Fashion-Gen using the following four fashion-related VL downstream tasks.

\noindent{\textbf{Task 1: Text-Image Retrieval (TIR).}}
The TIR task requires the model to find a text with the highest similarity value with different query images. In particular, we take a product title and its corresponding image as a positive image-text pair, while the negative pairs are randomly selected from a pool of mismatched images. To increase our experiment's difficulty, we constrain a set of image-text candidates (\ie, a positive pair and 100 negative pairs) in the same sub-category, making them as similar as possible.

\noindent{\textbf{Task 2: Image-Text Retrieval (ITR).}}
As the reverse process of the TIR task, the ITR task aims to retrieve a matching image given a sequence of text entries of fashion description, where these bidirectional retrieval tasks (\ie, TIR and ITR) become a prominent member of cross-modal research. Similar to the above selection strategy in the TIR, we prepare a set of candidate image-text pairs, including a positive pair and 100 negative pairs from the same sub-category. We evaluate the zero-shot learning ability of our \ourmodel~without further fine-tuning for these two retrieval tasks. We utilize three accuracy metrics (\ie, $\mathcal{R}$@1, $\mathcal{R}$@5, and $\mathcal{R}$@10) for the evaluation by ranking a series of predicted probabilities.

\noindent{\textbf{Task 3: Category Recognition (M-CR and S-CR).}}
This task has two parts: main-category recognition (M-CR) and sub-category recognition (S-CR).
These tasks act as the fundamental role of practical e-commerce applications that offer the specific category of the queried product. We expect that the model should possess the ability to recognize differences under different granularity levels: $48$ main-categories and $122$ sub-categories, such as $\{\text{M-CR}=\texttt{SWEATERS},\text{S-CR}=\texttt{CREWNECKS} \}$. After the class embedding in the last language embedding $\mathbf{T}^4$, we add two independent FC layers to generate the final probabilities for two different recognition tasks. This procedure requires additional fine-tuning with recognition labels. We utilize two recognition-related metrics to evaluate performance: accuracy ($\mathcal{A}$) and macro F-measure (macro-$\mathcal{F}$).

\noindent{\textbf{Task 4: Masked Image Generation (MIG).}}
MIG task can be viewed as a pixel-wise reconstruction task. Each patch in the image is randomly masked with the probability $r_v$ (refer to the pre-training task MIR in~\secref{sec:pre_training_loss}). Then, we ask the model to recreate the whole image using the uncovered areas as visual clues.

\begin{figure*}[t!]
  \centering
  \includegraphics[width=.98\linewidth]{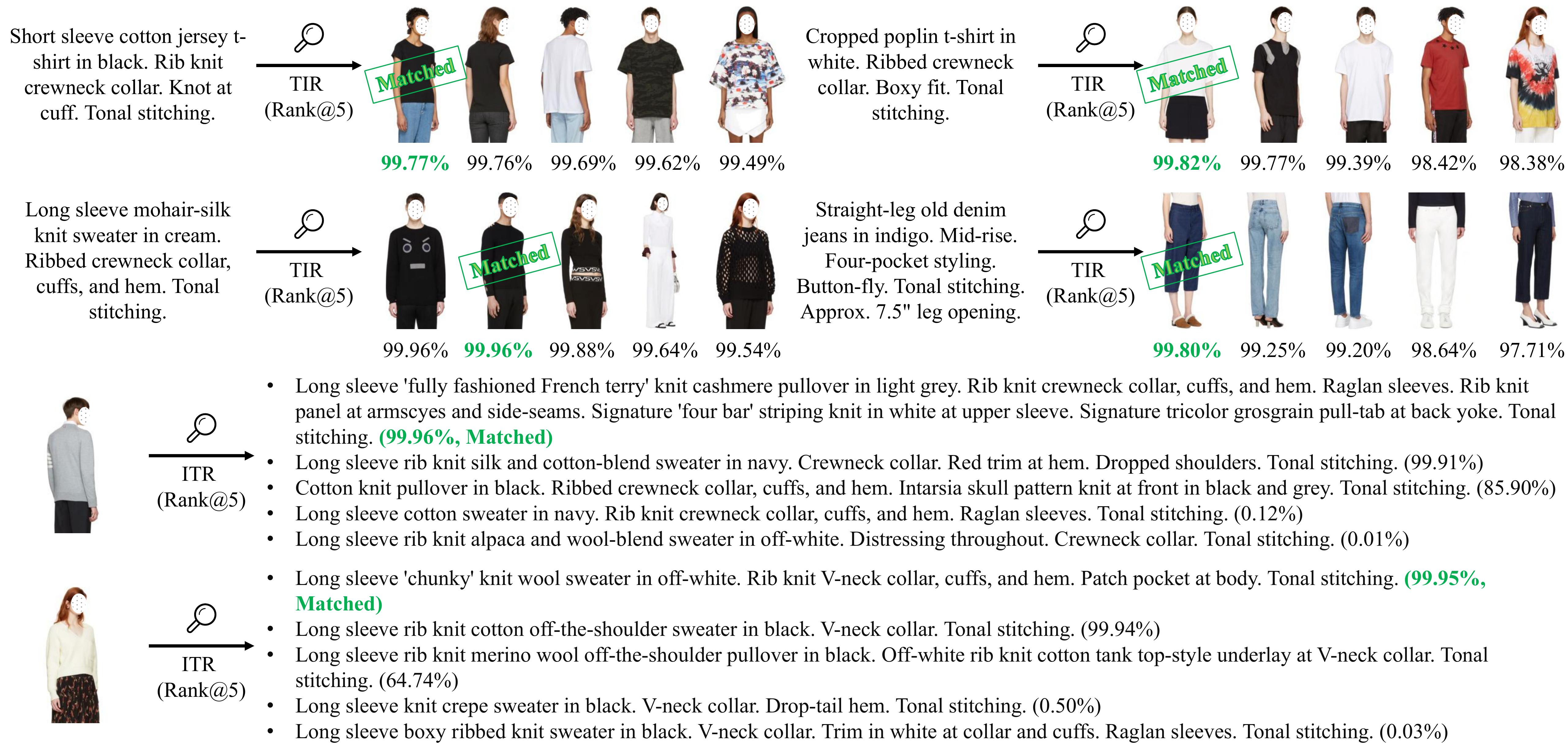}
  \caption{Visualization results on the TIR and ITR tasks in terms of top-five ranked probabilities predicted by our \ourmodel. ``Matched'' indicates the ground-truth image-text pair.
    }\label{fig:Re-viz}
\end{figure*}

%%%%%%%%%%%%%%%%%%  
\section{Experiments}\label{sec:Experiment}
%%%%%%%%%%%%%%%%%%  
This section will detail our experiment to determine the factors leading to the success of the proposed \ourmodel.

%%%%%%%%%%%%%%%%%%  
\subsection{Settings}
%%%%%%%%%%%%%%%%%%  
This part provides the hyperparameter settings for our training procedure: \textbf{i) Pre-training.} We utilize PyTorch to implement our method, which is accelerated by 8 Tesla V100 GPUs. We adopt AdamW optimizer with a momentum value of $0.9$, a mini-batch size of $1200$ (\ie, $150$ per GPU), a weight decay of $10^{-4}$. To avoid over-fitting, we initialize \ourmodel~on ImageNet pre-trained weights~\cite{wang2021pyramid}. The learning rate is initially set to $2.5\times10^{-3}$ and is changed using a cosine learning schedule. For the visual side, the input image is resized to $H$$=$$W$$=$$256$ and split into the multiple sub-patches with a size of $P=4$. For the language side, all the product captions are tokenized and padded to tokens with a unified length of $L=128$, including classification, caption, and padding tokens. The mask probabilities for vision and language are set to $r_v=0.5$ and $r_l=0.15$, respectively. We empirically set weighting factors $\{w_1=10,w_2=1,w_3=1\}$ to balance the orders of magnitude of different loss values. \textbf{ii) Fine-tuning.} We transfer the pre-trained VL representation to each downstream application via fine-tuning in an end-to-end manner, whose settings are consistent with the pre-training process.

\begin{figure*}[t!]
  \centering
  \includegraphics[width=\linewidth]{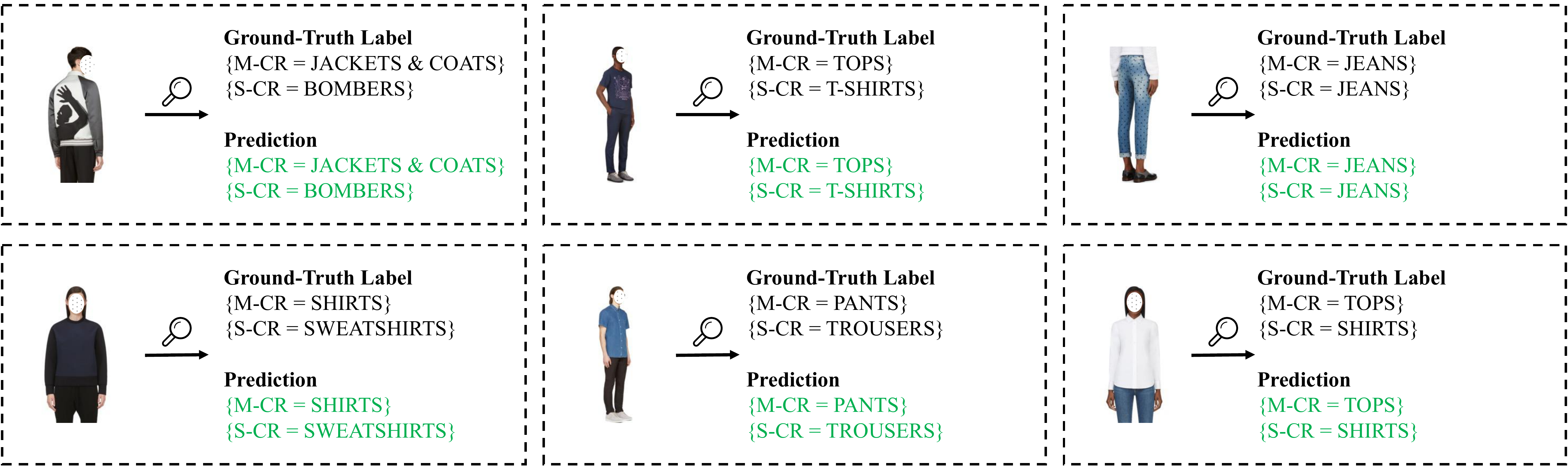}
  \caption{The visualization of main-/sub-category recognition results on Fashion-Gen. The green predictions hit the targets.
  }\label{fig:rec_viz_more}
\end{figure*}

\begin{figure*}[t!]
  \centering
  \includegraphics[width=\linewidth]{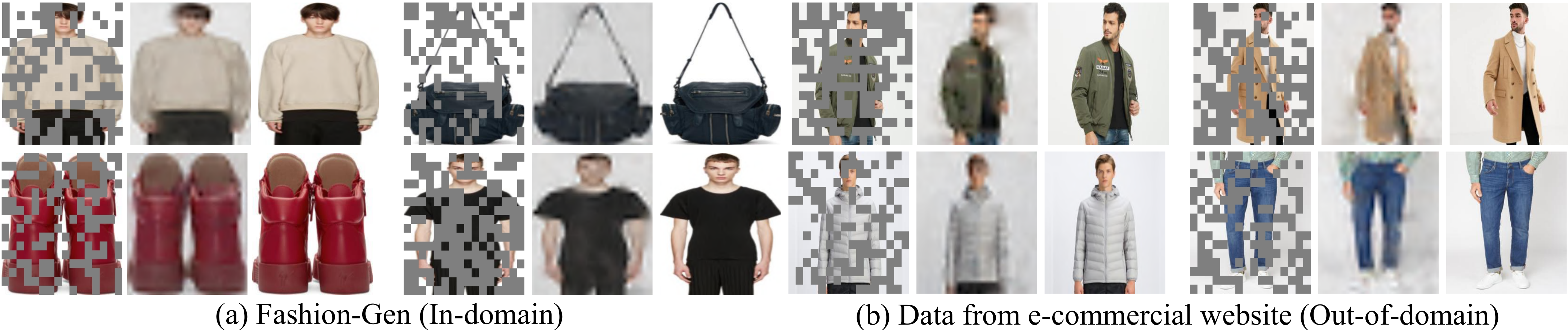}
  \caption{
    Visualization of samples generated by our \ourmodel. The gray blocks represent the masked regions.
  }\label{fig:mask_viz}
\end{figure*}

%%%%%%%%%%%%%%%%%%  
\subsection{Results}
%%%%%%%%%%%%%%%%%%  
As described in~\secref{sec:downstream_app}, we provide the details of four downstream fashion-related tasks. Experimental results show that our \ourmodel~outperforms all competitors, including VSE~\cite{kiros2014unifying}, VSE++~\cite{faghri2017vse++}, SCAN~\cite{lee2018stacked}, PFAN~\cite{wang2019position}, ViLBERT~\cite{lu2019vilbert}, ImageBERT~\cite{qi2020imagebert}, FashionBERT~\cite{gao2020fashionbert}, VL-BERT~\cite{su2020vl}, OSCAR~\cite{li2020oscar}, and Kaleido-BERT~\cite{Zhuge2021KaleidoBERT}, which demonstrate the superiority for handling the VL understanding and generation tasks.

\noindent \textbf{TIR and ITR.} 
As shown in~\tabref{tab:exp_i2t_t2i}, our \ourmodel~surpass the best method (\ie, Kaleido-BERT-CVPR$_{21}$) on the TIR task by margins of  $\textbf{+17.40\%}$, $\textbf{+20.91\%}$ across the $\mathcal{R}@5$, $\mathcal{R}@10$. As for ITR, our method delivers more competitive results, with improvements of $\textbf{+17.11\%}$, $\textbf{+22.73\%}$ on the  $\mathcal{R}@5$,  $\mathcal{R}@10$ metrics, respectively. In any case, these results strongly support that our model is powerful enough to match vision and language. They also show how \textbf{a) MIR} and \textbf{b) end-to-end pre-training} are useful in fashion. We believe that \ourmodel~would set a precedent in many industrial applications because it is a simple, cost-effective, and powerful architecture. Besides, we present the visualization results of these two retrieval tasks in~\figref{fig:Re-viz}.

\noindent \textbf{M-CR and S-CR.} 
Compared with BERT-based architectures~\cite{qi2020imagebert,gao2020fashionbert,li2020oscar,Zhuge2021KaleidoBERT}, we also achieve top-1 performances in these two tasks, demonstrating our method have an excellent VL understanding capability. Moreover, compared with the best method Kaleido-BERT, our architecture improves by $\textbf{0.193}$ in macro-$\mathcal{F}$ metric for the S-CR task. In addition, the mean improvements in terms of the Sum$\mathcal{C}$ metric (\ie, M-CR: $\textbf{+21.39}$ and S-CR: $\textbf{+24.80}$) are very significant. Since this metric is very sensitive to data distribution, it demonstrates \ourmodel~has super-strong robustness. We also present the recognition results of M-CR and S-CR in~\figref{fig:rec_viz_more}.

\noindent\textbf{MIG.} 
As shown in \figref{fig:mask_viz}, we showcase reconstructed images on the validation part of Fashion-Gen 2018 (a) and our e-commercial website (b). As seen, the reconstruction performance is truly remarkable. Since it requires our method to learn the fashion semantics truly, such results demonstrate the generative ability of our approach.

%%%%%%%%%%%%%%%%%%  
\subsection{Ablation Studies}\label{sec:ablation}
%%%%%%%%%%%%%%%%%%  
\textbf{Mask Ratio.} \tabref{tab:complete_ablation} (a) present four variants for different mask probability $r_v$ (\ie, $0.10$ (A1), $0.30$ (A2), $0.70$ (A3), $0.90$ (A4)) and our choice: $0.50$ \textbf{(Final)}. The  $\mathcal{R}@5$ rises steadily with the masking probability until it reaches the sweet spot ($75.70\% \rightarrow 78.00$\%); then it reach performance plummets ($73.80$\%). We argue that increasing the $r_v$ will make MIR more complex, allowing \ourmodel~to learn better semantics in a more restricted situation. However, masking out too much region will naturally result in losing valid visual information, leading to bad results.

\begin{table*}[t!]
\centering
\tiny
\renewcommand\arraystretch{1.5} 
\setlength\tabcolsep{0.5pt}
\caption{Ablation studies of five key pre-training factors on our \ourmodel. More relevant analyses refer to~\secref{sec:ablation}.
}\label{tab:complete_ablation}
    \begin{tabular}{l | l || cccc | cccc | ccc | ccc | c | c }
        \hline
        & &\multicolumn{4}{c|}{\textbf{(a) Mask Ratio ($r_v$)}} &\multicolumn{4}{c|}{\textbf{(b) Masking Unit Size ($\alpha$)}} &\multicolumn{3}{c|}{\textbf{(c) Masking Style}} &\multicolumn{3}{c|}{\textbf{(d) Pre-Training Tasks}} & \textbf{(e) Pre-Train}  & \textbf{\ourmodel} \\
        & & (A1)  & (A2) & (A3)  & (A4)  
        & (B1) & (B2) & (B3) & (B4)
        & (C1) & (C2) & (C3) 
        & (D1) & (D2) & (D3)
        & (E1)
        & \multirow{2}{*}{(\textbf{Final})} \\
        App. &Metric
        & $0.10$ & $0.30$ & $0.70$ & $0.90$
        & $1$ & $2$ & $8$ & $16$
        & Grid & Stroke & Center
        & ITM & ITM+MIR & ITM+MLM 
        & \textit{w/o} PVT
        &  \\
        \hline
         \multirow{3}{*}{TIR}
         & $\mathcal{R}$@1
         & 31.10\% & 33.50\% & 30.50\% & 30.70\%  
         & 31.90\% & 30.30\% & 30.00\% & 32.20\% 
         & 32.20\% & 31.40\% & 30.40\% 
         & 30.40\% & 32.20\% & 32.90\% 
         & 29.00\% 
         & \textbf{34.60\%} \\
         & $\mathcal{R}$@5 
         & 75.70\% & 76.00\% & 75.50\% & 73.80\%  
         & 75.30\% & 75.60\% & 73.90\% & 76.90\%
         & 75.30\% & 76.10\% & 75.10\% 
         & 74.10\% & 76.00\% & 76.20\%
         & 72.20\% 
         & \textbf{78.00\%} \\
         & $\mathcal{R}$@10 
         & 88.60\% & 88.70\% & 88.00\% & 88.60\%  
         & 89.60\% & 88.60\% & 88.20\% & 88.60\% 
         & 88.50\% & 89.20\% & 87.20\%
         & 83.50\% & 87.20\% & 88.60\%
         & 86.60\%
         &\textbf{89.50\%} \\  
         \hdashline
         &$\text{Sum}\mathcal{R}$ 
         & 195.40 & 198.20 & 194.00 & 193.10  
         & 196.80 & 194.50 & 192.10 & 197.70   
         & 196.00 & 196.70 & 192.70 
         & 188.00 & 195.40 & 197.70 
         & 187.80  
         & \textbf{202.10} \\
         &Diff & \textcolor{myGray}{$\scriptscriptstyle -6.70$} & \textcolor{myGray}{$\scriptscriptstyle -3.90$} & \textcolor{myGray}{$\scriptscriptstyle -8.10$} & \textcolor{myGray}{$\scriptscriptstyle -9.00$} & \textcolor{myGray}{$\scriptscriptstyle -5.30$} & \textcolor{myGray}{$\scriptscriptstyle -7.60$} & \textcolor{myGray}{$\scriptscriptstyle -10.00$} & \textcolor{myGray}{$\scriptscriptstyle -4.40$} & \textcolor{myGray}{$\scriptscriptstyle -6.10$} & \textcolor{myGray}{$\scriptscriptstyle -5.40$} & \textcolor{myGray}{$\scriptscriptstyle -9.40$} & \textcolor{myGray}{$\scriptscriptstyle -14.10$} & \textcolor{myGray}{$\scriptscriptstyle -6.70$} & \textcolor{myGray}{$\scriptscriptstyle -4.40$} & \textcolor{myGray}{$\scriptscriptstyle -14.30$} & \textcolor{myGray}{$\scriptscriptstyle -$} \\
         \hline
         \multirow{3}{*}{ITR}
         & $\mathcal{R}$@1 
         & 30.00\% & 29.90\% & 29.90\% & 28.50\%  
         & 29.00\% & 29.70\% & 29.00\% & 28.90\% 
         & 31.40\% & 31.10\% & 30.10\% 
         & 29.30\% & 30.40\% & 28.40\%
         & 25.60\% 
         & \textbf{33.10\%} \\
         & $\mathcal{R}$@5 
         & 75.70\% & 74.90\% & 76.50\% & 75.00\%  
         & 76.90\% & 77.10\% & 74.20\% & 77.30\%  
         & 77.40\% & 74.50\% & 73.90\% 
         & 70.80\% & 75.50\% & 76.30\%
         & 71.50\% 
         &\textbf{77.20\%} \\
         & $\mathcal{R}$@10 
         & 88.80\% & 89.00\% & 89.20\% & 88.20\%  
         & 89.40\% & 87.70\% & 88.00\% & 89.90\%  
         & 89.60\% & 88.50\% & 87.80\% 
         & 86.80\% & 87.80\% & 88.80\%
         & 85.90\% 
         & \textbf{91.10\%} \\
         \hdashline 
         &$\text{Sum}\mathcal{R}$  & 194.50 & 193.80 & 195.60 & 191.70  
         & 195.30 & 194.50 & 191.20 & 196.10   
         & 198.40 & 194.10 & 191.80
         & 186.90 & 193.70 & 193.50
         & 183.00 
         & \textbf{201.40} \\
         &Diff  & \textcolor{myGray}{$\scriptscriptstyle -6.90$} & \textcolor{myGray}{$\scriptscriptstyle -7.60$} & \textcolor{myGray}{$\scriptscriptstyle -5.80$} & \textcolor{myGray}{$\scriptscriptstyle -9.70$} & \textcolor{myGray}{$\scriptscriptstyle -6.10$} & \textcolor{myGray}{$\scriptscriptstyle -6.90$} & \textcolor{myGray}{$\scriptscriptstyle -10.20$} & \textcolor{myGray}{$\scriptscriptstyle -5.30$} & \textcolor{myGray}{$\scriptscriptstyle -3.00$} & \textcolor{myGray}{$\scriptscriptstyle -7.30$} & \textcolor{myGray}{$\scriptscriptstyle -9.60$} & \textcolor{myGray}{$\scriptscriptstyle -14.50$} & \textcolor{myGray}{$\scriptscriptstyle -7.70$} & \textcolor{myGray}{$\scriptscriptstyle -7.90$} & \textcolor{myGray}{$\scriptscriptstyle -18.40$} 
         & \textcolor{myGray}{$\scriptscriptstyle -$} \\
         \hline
         \multirow{2}{*}{M-CR}
         & $\mathcal{A}$ 
         & 98.16\% & 97.87\% & 98.09\% & 98.06\% 
         & 98.03\% & 98.04\% & 98.11\% & 98.01\%
         & 98.12\% & 98.07\% & 98.04\% 
         & 96.49\% & 97.11\% & 98.08\% 
         & 97.92\% 
         & \textbf{98.26\%} \\
         & macro-$\mathcal{F}$ 					
         & 0.870 & 0.860 & 0.890 & 0.870 
         & 0.870 & 0.880 & 0.850 & 0.870
         & 0.869 & 0.877 & 0.870 
         & 0.806 & 0.853 & 0.876 
         & 0.879 
         &\textbf{0.896}  \\
         \hdashline
         &$\text{Sum}\mathcal{C}$ 					
         & 185.16 & 183.87 & 187.09 & 185.06
         & 185.03 & 186.04 & 183.11 & 185.01
         & 185.02 & 185.77 & 185.04 
         & 177.09 & 182.41 & 185.68 
         & 185.82  
         & \textbf{187.86} \\
         &Diff					
         & \textcolor{myGray}{$\scriptscriptstyle -2.70$} & \textcolor{myGray}{$\scriptscriptstyle -3.99$} & \textcolor{myGray}{$\scriptscriptstyle -0.77$} & \textcolor{myGray}{$\scriptscriptstyle -2.80$} 
         & \textcolor{myGray}{$\scriptscriptstyle -2.83$} & \textcolor{myGray}{$\scriptscriptstyle -1.82$} & \textcolor{myGray}{$\scriptscriptstyle -4.75$} & \textcolor{myGray}{$\scriptscriptstyle -2.85$} 
         & \textcolor{myGray}{$\scriptscriptstyle -2.84$} & \textcolor{myGray}{$\scriptscriptstyle -2.09$} & \textcolor{myGray}{$\scriptscriptstyle -2.82$} 
         & \textcolor{myGray}{$\scriptscriptstyle -10.77$} & \textcolor{myGray}{$\scriptscriptstyle -5.45$} & \textcolor{myGray}{$\scriptscriptstyle -2.18$}  
         & \textcolor{myGray}{$\scriptscriptstyle -2.04$} 
         & \textcolor{myGray}{$\scriptscriptstyle -$} \\
         \hline
         \multirow{2}{*}{S-CR}
         & $\mathcal{A}$ 					
         & 93.10\% & 93.34\% & 93.36\% & 93.23\% 
         & 93.29\% & 93.34\% & 93.32\% & 93.32\% 
         & 93.37\% & 93.21\% & 93.59\% 
         & 89.64\% & 90.87\% & 93.29\%  
         & 92.90\% 
         & \textbf{93.57\%} \\
         & macro-$\mathcal{F}$ 					
         & 0.800 & 0.810 & 0.820 & 0.810 
         & 0.810 & 0.810 & 0.800 & 0.799
         & 0.794 & 0.814 & 0.830
         & 0.703 & 0.728 & 0.809 
         & 0.790
         &\textbf{0.829}  \\
         \hdashline	
         &$\text{Sum}\mathcal{C}$ 
         & 173.10 & 174.34 & 175.36 & 174.23 
         & 174.29 & 174.34 & 173.32 & 173.22
         & 172.77 & 174.61 & 176.59  
         & 159.94 & 163.67 & 174.19 
         & 171.90 
         & \textbf{176.47} \\
         &Diff 					
         & \textcolor{myGray}{$\scriptscriptstyle -3.37$} & \textcolor{myGray}{$\scriptscriptstyle -2.13$} & \textcolor{myGray}{$\scriptscriptstyle -1.11$} & \textcolor{myGray}{$\scriptscriptstyle -2.24$} & \textcolor{myGray}{$\scriptscriptstyle -2.18$} & \textcolor{myGray}{$\scriptscriptstyle -2.13$} & \textcolor{myGray}{$\scriptscriptstyle -3.15$} & \textcolor{myGray}{$\scriptscriptstyle -3.25$} & \textcolor{myGray}{$\scriptscriptstyle -3.70$} & \textcolor{myGray}{$\scriptscriptstyle -1.86$} & \textcolor{myGray}{$\scriptscriptstyle +0.12$}  & \textcolor{myGray}{$\scriptscriptstyle -16.53$} & \textcolor{myGray}{$\scriptscriptstyle -12.80$} & \textcolor{myGray}{$\scriptscriptstyle -2.28$}  & \textcolor{myGray}{$\scriptscriptstyle -4.57$} & \textcolor{myGray}{$\scriptscriptstyle -$} \\
    \hline
    \end{tabular}
\end{table*}

\noindent\textbf{Masked Unit Size.} 
Thanks to PVT's flexibility, we can easily try different sizes of masked patches. As shown in \tabref{tab:complete_ablation} (b), we derive four variants with masked unit size $\alpha$ (\ie, 1 (B1), 2 (B2), 8 (B3), 16 (B4)) to compare with our setting: 4 \textbf{(Final)}. We found the performance is sensitive to this factor. It makes sense, revealing how vital it is to learn a robust fashion-related representation with a moderate granularity.

\begin{figure}[t!]
  \centering
  \includegraphics[width=\linewidth]{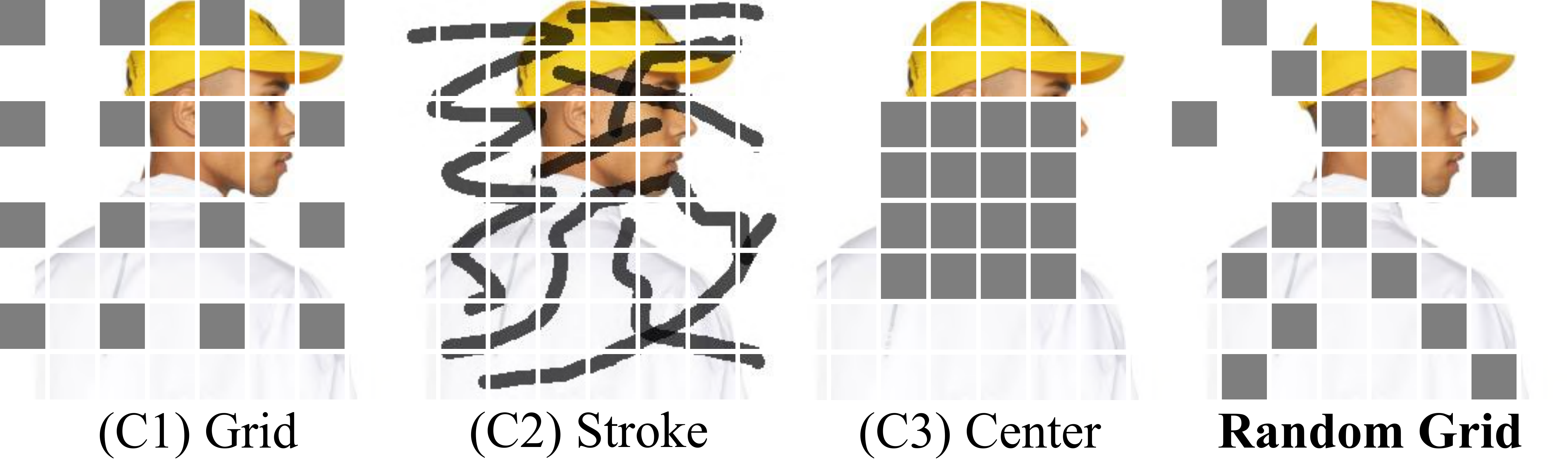}
  \caption{
    We designed four strategies to mask fashion images.  
    The random grid performs the best.
  }\label{fig:mask_types}
\end{figure}

\noindent\textbf{Masking Style.} 
As shown in \figref{fig:mask_types}, we designed four types of masking strategies for the MIR task, whose quantitative differences are presented in \tabref{tab:complete_ablation} (c), \ie, grid (C1), stroke (C2), center (C3) and our random grid (\textbf{Final}) masking strategies. As can be seen, the random grid masking (Final) yields the best results, while the other three perform poorly. 
We believe this is because, in comparison to the grid (C1) and center (C3), random grid masking (Final) can help \ourmodel~construct comprehensive representations. As our strategy (Final) does, the stroke (C2) also randomly masks the image given, yet it more or less leaves unmasked visual cues in the sub-patches. Our strategy enables the model to easily predict the masked region because semantics in the image are well preserved, enhancing the model's robustness to learning in-sight knowledge.

\noindent\textbf{Pre-Training Objectives.}
As shown in \tabref{tab:complete_ablation} (d), we derive four different variants to investigate the contribution of each objective, including ITM (D1), ITM+MIR (D2), ITM+MLM (D3), and our ITM+MIR+MLM \textbf{(Final)}. When comparing D3 to D1 and D2 in the TIR task, we can see that D3 has a better performance in $\mathcal{R}@5$ metric: $74.10$\% (D1)~\textless~$76.00$\% (D2)~\textless~$76.20$\% (D3). We conclude MLM task can help the model thoroughly learn the language knowledge, so it provides a more precise query to recall better-matching images. In the ITR task, we find the similar conclusion when comparing (D2) to (D1) and D3 in $\mathcal{R}@5$ metric: $70.80$\% (D1)~\textless~$75.50$\% (D2)~\textless~$76.30$\% (D3). It indicates that better visual learning leads to an accurate image query to match the most appropriate caption.

\noindent\textbf{Loading Pre-Trained Weight.} 
As seen in \tabref{tab:ablation_results}, we add an experiment to demonstrate it is very important to load the PVT's weight pre-trained on ImageNet~\cite{deng2009imagenet}. If not, it is obvious that our \ourmodel~will suffer fierce drops (\ie, ITR: $77.20 \% \rightarrow 71.50 \%$ in $\mathcal{R}@5$, S-CR:  $93.57\% \rightarrow 92.90\%$ in $\mathcal{A}$). It is reasonable because a method pre-trained on large-scale general datasets can be more applicable in a specific field. It has already learned information such as color, texture, shape, \etc.

\begin{table}[t!]
\centering
\tiny
\caption{Ablation study for the contribution of loading PVT's weights pre-trained on ImageNet~\cite{deng2009imagenet}.}
\label{tab:ablation_results}
\renewcommand{\arraystretch}{1.3}
\renewcommand{\tabcolsep}{0mm}
\begin{tabular}{r|cc|cc|cc|cc}
    \hline
    &\multicolumn{2}{c|}{TIR} &\multicolumn{2}{c|}{ITR} &\multicolumn{2}{c|}{M-CR} &\multicolumn{2}{c}{S-CR}\\
    % \cline{2-9}
     &$\mathcal{R}@5$ &$\mathcal{R}@10$ &$\mathcal{R}@5$ &$\mathcal{R}@10$
     &$\mathcal{A}$ &macro-$\mathcal{F}$ &$\mathcal{A}$ &macro-$\mathcal{F}$ \\
    \hline
    \textit{w/o} PVT
    &72.20\% &86.60\% &71.50\% &85.90\% &97.92\% &0.879 &92.90\% &0.790 \\
    \textit{w/} PVT 
    &\textbf{78.00\%} &\textbf{89.50\%} &\textbf{77.20\%} &\textbf{91.10\%}&\textbf{98.26\%} &\textbf{0.896} &\textbf{93.57\%} &\textbf{0.829} \\
    Diff & \textcolor{myGray}{ $\scriptscriptstyle +5.80\%$ } &\textcolor{myGray}{$\scriptscriptstyle +2.90\%$} &\textcolor{myGray}{$\scriptscriptstyle +5.70\%$} &\textcolor{myGray}{$\scriptscriptstyle +5.20\%$} &\textcolor{myGray}{$\scriptscriptstyle +0.34\%$} &\textcolor{myGray}{$\scriptscriptstyle +1.7\%$} &\textcolor{myGray}{$\scriptscriptstyle +0.67\%$} &\textcolor{myGray}{$\scriptscriptstyle +3.9\%$} \\
    \hline
\end{tabular}
\end{table}

%%%%%%%%%%%%%%%%%%  
\subsection{More Discussions}\label{sec:discussion}
%%%%%%%%%%%%%%%%%%  
\noindent \textbf{How does MVLT perform in general domains?} To further investigate the potential abilities in general settings, we here discuss two extended questions. \textit{a) Can the general models be directly transferred to the fashion domain?} Inspired by the huge impact of general vision-language models, as in~\tabref{tab:clip}, we further investigate the zero-shot performance of two typical general models (\ie, ViLBERT~\cite{lu2019vilbert} and CLIP~\cite{radford2021learning}). This has once again demonstrated the necessity and superiority of \ourmodel~pre-trained on the specific domains. \textit{b) Can \ourmodel~also work well in the general domain?} We further verify the potential ability of our \ourmodel~on the general domain. \tabref{tab:mscoco} reports the performance on MS-COCO 2014 dataset~\cite{lin2014microsoft}, where \ourmodel~follows the same training standards as in~\cite{kim2021vilt}. It shows that \ourmodel~achieves promising results compared to the latest models (\ie, Unicoder-VL~\cite{li2020unicoder}, UNITER~\cite{chen2020uniter}, and ViLT~\cite{kim2021vilt}) without extra training data and special retrieval losses during the training. It indicates that \ourmodel~is also a promising solution when extended to general scenes.

\begin{table}[t!]
\centering
\tiny
\renewcommand\arraystretch{1.5}
\setlength\tabcolsep{0.3pt}
\caption{The comparison of zero-shot retrieval results on the Fashion-Gen dataset.}\label{tab:clip}
    \begin{tabular}{ c || ccc | ccc }
    \hline
    &\multicolumn{3}{c|}{TIR} &\multicolumn{3}{c}{ITR } \\
    &$\mathcal{R}$@1$\uparrow$ &$\mathcal{R}$@5$\uparrow$ &$\mathcal{R}$@10$\uparrow$ &$\mathcal{R}$@1$\uparrow$ &$\mathcal{R}$@5$\uparrow$ &$\mathcal{R}$@10$\uparrow$ \\
    \hline
    ViLBERT (Zero-shot) &7.18\% &18.73\% &29.84\% &8.99\% &15.34\% &26.14\% \\
    CLIP (Zero-shot) &16.30\% &40.60\% &55.60\% &13.60\% &43.10\% &57.60\% \\
    \hline
    \textbf{\ourmodel~(OUR)} &\textbf{34.60\%} &\textbf{78.00\%} &\textbf{89.50\%} &\textbf{33.10\%} &\textbf{77.20\%} &\textbf{91.10\%} \\
    \hline
    \end{tabular}
\end{table}

\begin{table}[t!]
\centering
\tiny
\renewcommand\arraystretch{1.5}
\setlength\tabcolsep{1.8pt}
\caption{Retrieval results on the MS-COCO 2014 dataset. $\dag$ means using an extra feature extractor (\eg, Faster RCNN).}\label{tab:mscoco}
    \begin{tabular}{ c || ccc| ccc }
    \hline
&\multicolumn{3}{c|}{TIR task (5K Test)} &\multicolumn{3}{c}{ITR task (5K Test)} \\
    &$\mathcal{R}$@1$\uparrow$ &$\mathcal{R}$@5$\uparrow$ &$\mathcal{R}$@10$\uparrow$ &$\mathcal{R}$@1$\uparrow$ &$\mathcal{R}$@5$\uparrow$ &$\mathcal{R}$@10$\uparrow$ \\
    \hline
Unicoder-VL$^\dag$ &48.40\% &76.70\% &85.90\% &62.30\% &87.10\% &92.80\% \\
UNITER-Base$^\dag$ &\textbf{50.30\%} &78.50\% &87.20\% &64.40\%  &87.40\%  &93.10\% \\
ViLT-Base/32 &41.30\% &72.00\% &82.50\% &61.80\% &86.20\% &92.60\% \\
\hline
\textbf{\ourmodel~(OUR)} &49.66\% &\textbf{79.88\%} &\textbf{87.50\%} &\textbf{65.38\%} &\textbf{90.04\%} &\textbf{93.60\%}\\
    \hline
    \end{tabular}
\end{table}

\noindent \textbf{Why do pyramid architecture and MIR benefit?} As mentioned in the introduction, there are two understudied problems in the fashion domain. \emph{To solve the transferability problem}, pyramidal architecture~\cite{wang2021pyramid} takes raw data as input without complex pre-processing, which essentially alleviates the applied burden in industry. Besides, MIR does not need human annotations like classification tags, bounding boxes, or pixel-wise segmentation labels. \emph{For the granularity problem}~\cite{wu2022multi}, the pyramidal architecture~\cite{wang2021pyramid} provides multi-scale features with rich semantics. Combined with the MIR task, our framework can represent multi-grained fashion knowledge (\eg, dress, V-neck). These features are helpful and urgently required in this field.

A VL model that performs well for semantic understanding tasks (\eg, retrieval~\cite{chen2022cross}, classification) can serve as a good foundation and be easily applied to downstream tasks (\eg, text-to-image synthesis~\cite{liu2022verbal}, image captioning) by utilizing an additional decoder. We didn't conduct image captioning experiments  because we focused on basic representation learning in fashion this time.

\noindent\textbf{MVLT \emph{v.s.} MAE~\cite{he2021masked}.} 
MAE learns general representations by allowing the model to explore pixel-to-pixel associations. So \ourmodel~and MAE are similar in this regard. However, our \ourmodel~is the first that introduces the vision reconstruction-alike pre-training for multi-modal research (\eg, fashion domain).

%%%%%%%%%%%%%%%%%%  
\section{Conclusion}
%%%%%%%%%%%%%%%%%%  
We present a vision-language framework named \ourmodel, which provides two contributions in this field: 1) a newly-designed masked image reconstruction (MIR) objective, and 2) an end-to-end pre-training scheme. 
The experimental and ablative analysis demonstrates the superiority of various matching and generative tasks.
\ourmodel~outperforms the cutting-edge method Kaleido-BERT with large margins on retrieval and recognition tasks, which would catalyze the fashion domain.
The designed out-of-box method working end-to-end could simplify the workflow (\eg, data pre-processing and model training) for the actual engineering value, which improves development and business efficiency on large-scale e-commerce websites by approximately 50\%.

In the future, we will continue to investigate an extremely efficient method in this field using famous technologies like hashing~\cite{zhang2022modality}, network pruning, and knowledge distil to alleviate the storage and computing limitations in real-world e-commerce applications.

%%%%%%%%%%%%%%%%%%  
\section*{Acknowledgments}
%%%%%%%%%%%%%%%%%%  
This work is funded by Toyota Motor Europe via the research project TRACE-Zürich. The authors also would like to thank the anonymous reviewers and editor for their helpful comments on this manuscript.

%%%%%%%%%%%%%%%%%%  
\section*{Conflicts of Interests}
%%%%%%%%%%%%%%%%%%  
The authors declared that they have no conflicts of interest to this work. We declare that we do not have any commercial or associative interest that represents a conflict of interest in connection with the work submitted.

\bibliographystyle{IEEEtran}
\bibliography{mir-article}

\end{document}